\documentclass[letterpaper]{article}

\usepackage{natbib,alifeconf}  

\usepackage{amsfonts}
\usepackage{amsmath}
\usepackage[ruled,vlined]{algorithm2e}
\usepackage{hyperref}

\title{Differentiable Programming of Reaction-Diffusion Patterns}
\author{Alexander Mordvintsev$^1$, Ettore Randazzo$^1$ 
\and Eyvind Niklasson$^1$ \\
\mbox{}\\
$^1$Google \\
moralex@google.com} 

\begin{document}
\maketitle

\begin{abstract}
Reaction-Diffusion (RD) systems provide a computational framework that governs many pattern formation processes in nature. Current RD system design practices boil down to trial-and-error parameter search. We propose a differentiable optimization method for learning the RD system parameters to perform example-based texture synthesis on a 2D plane. We do this by representing the RD system as a variant of Neural Cellular Automata and using task-specific differentiable loss functions. RD systems generated by our method exhibit robust, non-trivial ``life-like'' behavior.

\end{abstract}

\section{Introduction}

Multicellular organisms build and maintain their body structures through the distributed process of local interactions among tiny cells working toward a common global goal. This approach, often called \emph{self-organisation}, is drastically different from the way most human technologies work. We are just beginning to scratch the surface of integrating some of nature's ``best practices'' into technology.

In 1952, Alan Turin wrote his seminal work, ``The Chemical Basis of Morphogenesis'' \citep{Turing_1952}, in which he proposed that pattern formation in living organisms can be controlled by systems of chemical substances called \emph{morphogens}. Simultaneous processes of chemical Reactions-Diffusion (RD) of these substances provide the sufficient biocomputation platform to execute distributed algorithms of pattern formation \citep{Kondo2010ReactionDiffusionMA,Landge2019PatternFM}.

Even a system of just two interacting substances (e.g. Gray-Scott) can produce a great diversity of patterns and interesting behaviours \citep{Pearson_1993}. As is often the case with complex emergent behaviours of simple systems, it is very difficult find model parameters that produce a particular, predefined behaviour. Most of the time researchers use hand-tuned reaction rules, random and grid search over the parameter space for the combinations with the desired properties. Procedural texture synthesis is one of the best known applications for this type of parameter tuning \citep{Witkin1991ReactiondiffusionT,Turk1991GeneratingTO}. In this paper, we propose to use differentiable optimization as an alternative to such a trial-and-error process.

The task of determining the sets of parameters that lead to desired behaviors becomes even more important as we enter the realm of artificial life and synthetic biology. The work of \cite{Scalise2014DesigningMR} is a remarkable example of an attempt to design a flexible modular framework for RD-based spatial programming. The authors demonstrate (at least in simulation) a way to combine a set of basic computational primitives, implemented as RD systems of DNA strands, into multistage programs that generate non-trivial spatial structures. We argue that these programs cannot yet be called ``self-organizing systems'' due to two important limitations. First, they rely on a predefined initial state that defines the global coordinate system on which the program operates through chemical gradients or precise locations of chemical ``seeds". Second, program execution is a feed-forward process that does not imply homeostatic feedback loops that make the patterns robust to external perturbations or a changed initial state.

Another very related line of research in {\em artificial life} is Lenia \citep{Chan2020LeniaAE}, which aims to find rules and configurations for space-time-state-continuous cellular automata that demonstrate life-like homeostatic and replicating behaviours.

Figures in video form and a reference implementation is available here\footnote{\url{https://selforglive.github.io/alife_rd_textures/}}.

\section{Differentiable Reaction-Diffusion Model}
We study a computational model that can be defined by the following system of PDEs:

$$
\frac{\partial x_i }{\partial  t } = c_i\nabla^{2}x_i + f_\theta(x_0,\ldots, x_{n-1}) 
$$
$x_0,\ldots, x_{n-1}$ are scalar fields representing the ``concentrations'' $n$ ``chemicals'' on a 2D plane. $c_i$ are per-"chemical'' diffusion coefficients, and $f_\theta \colon\mathbb{R}^n \rightarrow \mathbb{R}^n$ is a function defining the rate of change of ``chemical concentrations'' due to local ``reactions". Chemical terms are used here in quotes because, in the current version of our model, the function $f_\theta$ need not obey any actual physical laws, such as the law of conservation of mass or the law of mass action. ``Concentrations'' can also become negative.

\subsection{Reaction-Diffusion as a Neural CA}
The objective of this paper is to train a model that can be represented by a space-time continuous PDE system. Yet, we have to discretize the model to run it on a digital computer. The discretized model can be treated as a case of Neural Cellular Automata (NCA) model \citep{mordvintsev2020growing, randazzo2020self-classifying}.

Our model and the training procedure are heavily inspired by the texture-synthesizing Neural CA described by \cite{niklasson2021self-organising}. Similarly, we discretize space into cells and time into steps, use explicit Euler integration of the system dynamics and backpropagation-through-time to optimize the model parameters. Our model differs from the previous texture-synthesizing NCA:

\begin{itemize}
\item CA iteration does not have the \textit{perception} phase. The ``reaction'' part of cell update (modelled by $f_\theta$) depends on the current state of the cell only.
\item There is an isotropic diffusion process that runs in parallel with ``reaction'' and is modelled by channelwise convolution with a Laplacian kernel. This is the only method of communication between cells. Thus, the Neural RD model is completely isotropic: in addition to translation, the system behaviour is invariant to rotational and mirroring transformations.
\item We do not use stochastic updates, all cells are updated at every step.
\end{itemize}

Thus, the system described here can be seen as a stripped version of the Neural CA model. These restrictions are motivated by a number of practical and philosophical arguments described below.

\paragraph{Model simplicity and prospects of physical implementation} The discussion section of ``Growing NCA'' article by \cite{mordvintsev2020growing} contains speculations about potential of physical implementation of the Neural CA as a grid of tiny independent computers (microcontrollers). Neural CA consists of a grid of discrete cells that are sophisticated enough to persist and update individual state vectors, and communicate within the neighborhood in a way that differentiates between neighbours, adjacent to different sides of the cell. This implies the existence of global alignment between cells, so that they agree where up and left are, and separate communication channels to send the information in different directions. In contrast, diffusion based communication doesn't require any dedicated structures, and ``just happends'' in nature due to the Brownian motion of molecules.

Furthermore, states of Neural CA cells are only modified by their learned update rules and are not affected by any environmental processes. Cells have clear separation between what's inside and outside of them, and can control which signals should be let through. However, many natural phenomena of self-organisation can be effectively described as a PDE system on a continuous domain. Individual elements that constitute the domain are considered negligibly small, and the notion of their individual updates is meaningless. 

In the experiments section we cover some practical advantages of the proposed RD model with respect to Neural CA. In particular we demonstrate the generalization into arbitrary mesh surfaces and even volumes.

\paragraph{Reaction-Diffusion CA update rule}
The discrete update rule can be written as follows:
\begin{equation}
\begin{gathered}
\label{eq:update}
x_{i}^{t+1} = x_i^{t} + c_i K_{lap} \ast x_i^t \, d + f_\theta(x_0^{t},\ldots, x_{n-1}^{t}) \, r \\
 d=\Delta_t / \Delta_{h}^2, \;\;   r=\Delta_t
\end{gathered}
\end{equation}
$K_{lap}$ is a 3x3 Laplacian convolution kernel, $c_i$ and $\theta$ are parameters that control the CA behaviour, and the coefficients $r$ and $d$  control the rates of reaction and diffusion, encapsulating temporal and spatial discretization step sizes $\Delta_t$ and $\Delta_h$. By varying these coefficients we can validate if the learned discrete CA rule approximates the continuous PDE and does not over-feat to the particular discretization. During training we use $r = d = 1.0$.

The function $f_\theta(\mathbf{x}) =  act(\mathbf{x} W_0  + b_0) W_1$ is a small two layer neural network with parameters $\theta: (W_0 \in \mathbb{R}^{n\times h}, W_1 \in \mathbb{R}^{h\times n}, b_0 \in \mathbb{R}^h)$ and a non-linear elementwise activation function $act$ (see the experiments section). In our experiments we use $n=32, h=128$, so the system models the dynamics of 32 ``chemicals'', and the total number of network parameters equals to 8320. Per-``chemical'' diffusion coefficients $c_i$ can be learned (in this case we set $c_i=sigmoid(\hat{c}_i)$ to make sure that diffusion rate stays in $0..1$ range), or fixed to specific values.

\section{Texture synthesis}
Reaction-Diffusion models are a well-known tool for texture synthesis. Typically, manual parameter tuning has been used to design RD systems that generate desired visual patterns \citep{Witkin1991ReactiondiffusionT,Turk1991GeneratingTO}. We propose an example-based training procedure to learn a RD rule for a given example texture image. This procedure closely follows the work ``Self-Organising Textures'' (SOT) \citep{niklasson2021self-organising}, with a few important modifications. Our goal is to learn a RD update function whose continuous application from a {\em starting state} would produce a pattern similar to the provided texture sample. The procedure is summarised in algorithm \ref{alg:tex}.

\begin{algorithm}
\label{alg:tex}
\SetAlgoLined
 \caption{Texture RD system training}
 $pool \leftarrow N_{pool}$ randomly sampled {\em seed states} \;
 \For{$step \leftarrow 1$ \KwTo $N_{train}$} {
    $batch \leftarrow$ sample $N_{batch}$ random $pool$ indices \;
    $x \leftarrow pool[batch]$\;
    \If {$step \, \bmod R_{seed} = 0$} {
        $x[0] \leftarrow$ random {\em  seed state}
    }
    \For{$i \leftarrow 1$ \KwTo $Uniform(I_{min}, I_{max})$} {
        $x \leftarrow RD_\theta(x)$ \;
    }
    $loss \leftarrow calcLoss(x)$ \;
    update $\theta$ using the gradient of $loss$ \;
    $pool[batch] \leftarrow $x\;
 }
\begin{tabular}{ | l r | l r | l r | }
\hline
 $N_{train}$  & 20000 & $I_{min}$ & 32  & $N_{pool}$ & 1024 \\
 $R_{seed}$   & 32    & $I_{max}$ & 96  & $N_{batch}$ & 4\\
\hline 
\end{tabular}
\end{algorithm}

\begin{figure}[h]
\includegraphics[width=8.5cm]{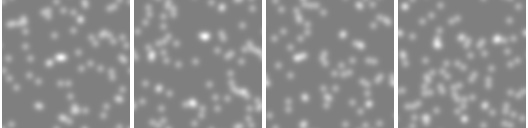}
\centering
\caption{Random {\em seed states}.}
\label{fig:seeds}
\end{figure}

\paragraph{Seed states} SOT uses seed states filled with zero values. Stochastic cell updates provide the sufficient variance to break the symmetry between cells. We use synchronous explicit Euler integration scheme that updates all cells simultaneously, so the non-uniformity of the seed states is required to break the symmetry. We initialize the grid with a number of sparsely scattered Gaussian blobs (fig.\ref{fig:seeds}).

Periodic injection of seed states into training batches is crucial to prevent the model from forgetting how to develop the pattern from the seed, rather then improving already existing one. We observed that it is sufficient to inject the seed much less often than in SOT. We use $R_{seed}  = 32$ in this experiment.

\begin{figure}
\includegraphics[width=8.5cm]{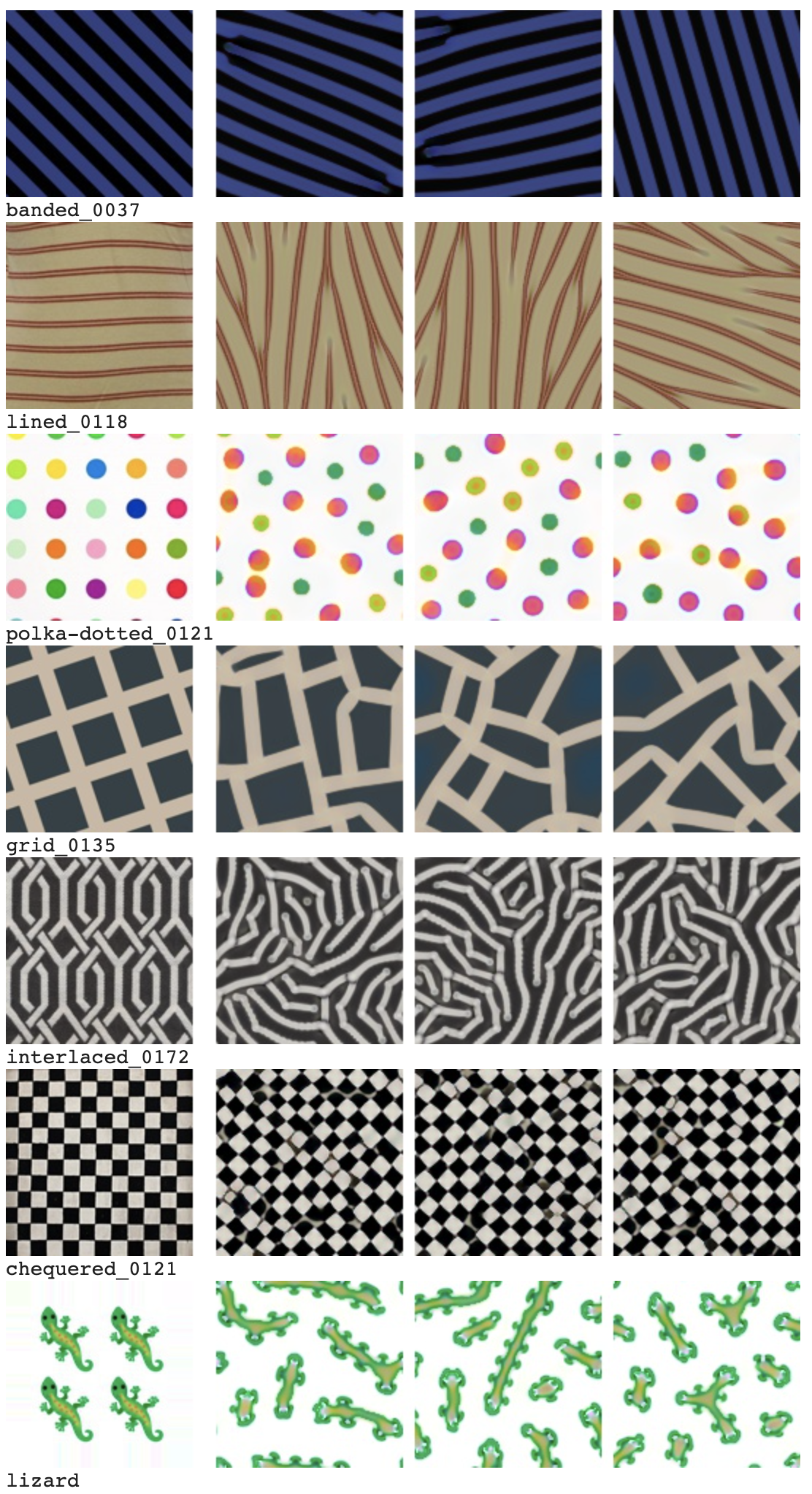}
\centering
\caption{Left column: sample texture images; other columns: different 5000-step runs of the learned RD models. Almost all models replicate features of the target texture in rotation-invariant fashion. Only {\tt chequered\_0121} overfitted to exploit the underlying raster grid structure to always produce diamond-oriented checker squares.}
\label{fig:textures}
\end{figure}

\paragraph{Rotation-invariant texture loss} Similar to SOT, we interpret the first 3 ``chemicals'' as RGB colors and use them to calculate texture loss. Our texture loss is also based on matching Gram matrices of pre-trained VGG network activations \citep{Gatys2015TextureSU}, but has an important modification to account for the rotational invariance of isotropic RD. Consider the texture {\tt lined\_0118} from the figure \ref{fig:textures}. The target pattern is anisotropic, but RD, unlike NCA, has no directional information and cannot produce a texture that exactly matches the target orientation. We construct the rotation-invariant texture loss by uniformly sampling $N_{rot}$ target images rotated by angles $0^\circ ... 360^\circ$ and computing the corresponding texture descriptions. This computation only occurs in the initialization phase and does not slow down the training. At each loss evaluation, the texture descriptors from RD are matched with all target orientations and the best match is taken for each sample in the batch.

\paragraph{} Per-``chemical'' diffusion coefficients in Eq. (\ref{eq:update}) are set to $c_{0..7}=1/8, c_{8..15}=1/4, c_{16..23}=1/2, c_{24..31}=1,  $  so that the ``substances'' are split into 4 groups of varying diffusivity. Therefore,  RGB colors correspond to slowly diffusing channels $x_{0..2}$. We experimented with learned diffusion coefficients, but it did not seem to bring substantial improvement, so we kept fixed values for the simplicity. In all of the texture synthesis experiments we use wrap-around (torus) grid boundary conditions.

RD network uses a variant of Swish \citep{Ramachandran2017SwishAS} elementwise activation function: $act(x) = x\sigma(5.0*x)$, where $\sigma$ is a sigmoid function.

We trained seven texture-synthesizing RD models (fig.\ref{fig:textures}). Six were using 128x128 images from DTD dataset \citep{Cimpoi2014DescribingTI}, and the last was using a 48x48 lizard emoji image, replicated four times. Models were trained for 20000 steps using the Adam optimiser \citep{Kingma2015AdamAM}  with learning rate 1e-3 decayed by 0.2 at steps 1000 and 10000. We also used the gradient normalization trick from SOT to improve training stability. Training a single RD model took about 1 hour on the NVIDIA P100 GPU.

\subsection{Results}

In spite of the constrained computational model, trained RD systems were capable of reproducing (although imperfectly) distinctive features of the target textures. All models except {\tt chequered\_0121} seemed to be isotropic, which manifested in the random orientation of the resulting patterns, depending on the randomized seed state.  {\tt chequered\_0121} always produced diamond-oriented squares, which suggests overfitting to the particular discrete grid. Below we investigate this behaviour more carefully.

\begin{figure}[h]
\includegraphics[width=8.5cm]{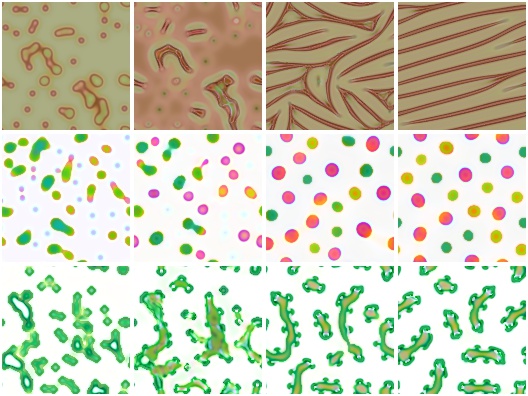}
\centering
\caption{Stages of texture development at time steps 50, 100, 1000 and 10000. Anisotropic textures form by local symmetry breaking and refinement. Resulting patterns look consistent over long time periods, but never reach full stability. For example, dots in the middle row asynchronously change colors.}
\label{fig:evol}
\end{figure}

\cite{Witkin1991ReactiondiffusionT} proposed using anisotropic diffusion for anisotropic texture generation with RD. In our experiments, we demonstrated the capability of fully isotropic RD systems to produce locally anisotropic textures through the process of iterative alignment, that looks surprisingly ``life-like". Figure \ref{fig:evol} shows snapshots of grid states at different stages of pattern evolution. We recommend watching supplementary videos to get a better intuition of RD system behaviours.

\paragraph{Do we really learn a PDE?}
We decided to validate that the discrete Neural CA that we use to simulate RD system is robust enough to be used with different grid resolution than used during training. This would confirm that the system we trained really approximates a space-time continuous PDE without overfitting to the particular grid discretization. One way to execute the RD on a finer grid is to decrease the $\Delta_h$ coefficient in (\ref{eq:update}). This leads to the quick growth of the diffusion term, making the simulation unstable. We can mitigate this by reducing the $\Delta_t$ as well. In practice we keep $d=1.0$, but decrease $r$, so that $\Delta_h = \sqrt{r}$. Thus, setting $r=1/4$ corresponds to running on a twice finer grid, and should produce patterns magnified two times. Decrease of $r$ may also be interpreted as reaction speed slowdown or diffusion acceleration.

\begin{figure}[t!]
\includegraphics[width=2.6cm]{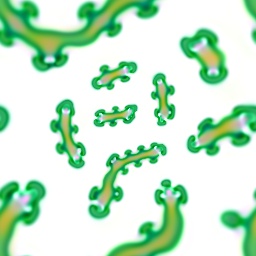}
\includegraphics[width=2.6cm]{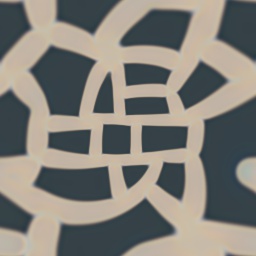}
\includegraphics[width=2.6cm]{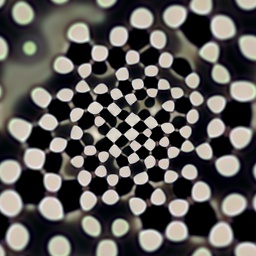}
\centering
\caption{Running RD system on a non-uniform $r$ grid. All models except the rightmost seem to be capable of operating on a modified grid, preserving the key pattern characteristics. {\tt chequered\_0121} overfitted to the particular grid resolution and is unable to produce right corners at larger scales. Sometimes it even develops instabilities and explodes.}
\label{fig:zoom}
\end{figure}

Fig. \ref{fig:zoom} shows results of RD system evaluation on a grid having non-uniform $r$: in the center $r=1$, slowly decreasing to $r \approx 1/9$ at the boundary. Most of the trained models were capable of preserving their behaviour independent of the grid resolution. {\tt chequered\_0121} model was the only exception. The grid-overfitting hypothesis was confirmed by the fact that the model could not form right angles at the grid corners, and even developed instabilities in the fine resolution grid areas.

\begin{figure}[t!]
\includegraphics[trim=100 20 0 0,width=4.0cm]{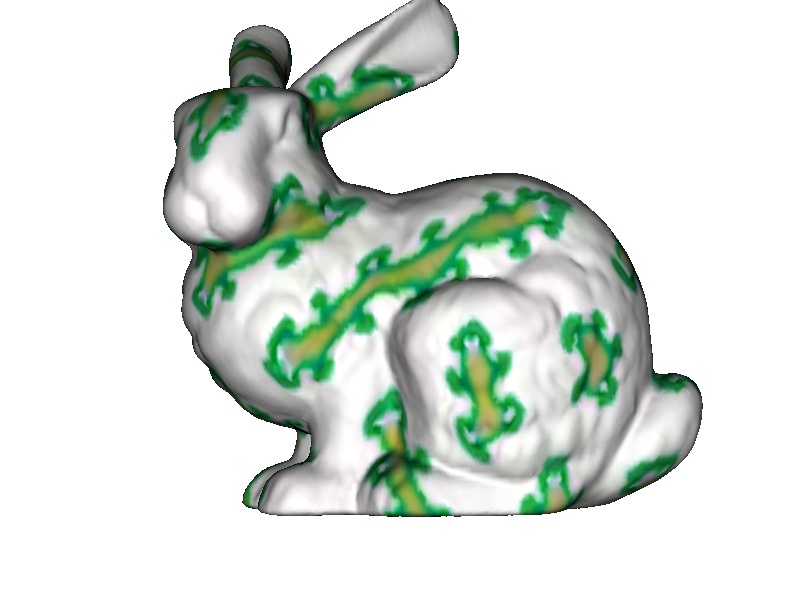}
\includegraphics[trim=100 20 0 0,width=4.0cm]{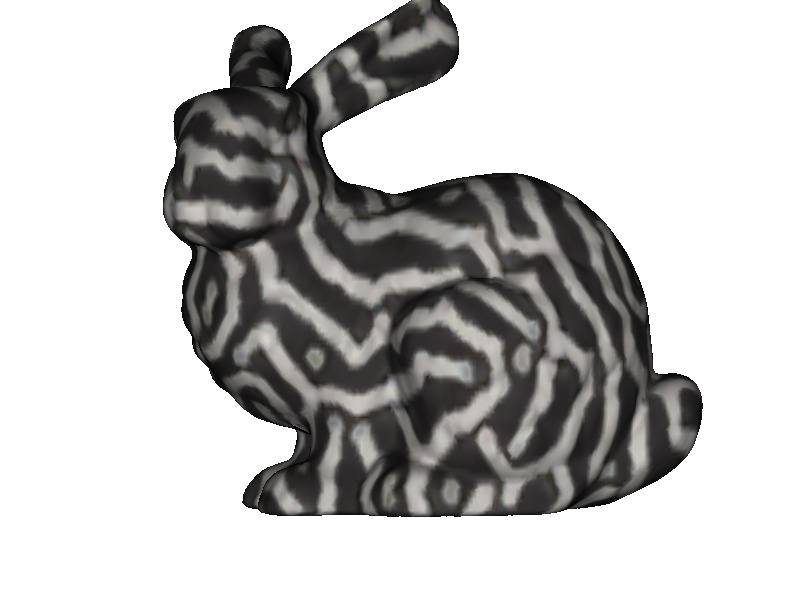}
\includegraphics[trim=100 20 0 0,width=4.0cm]{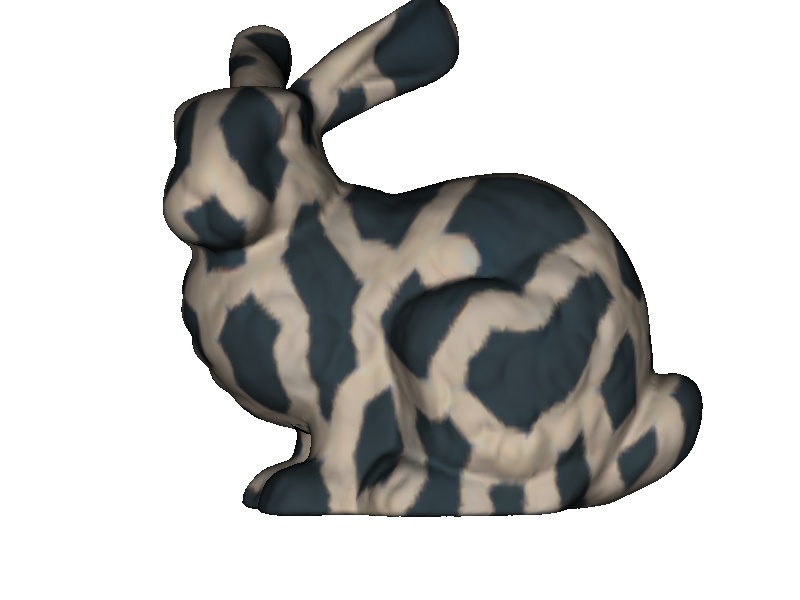}
\includegraphics[trim=100 20 0 0,width=4.0cm]{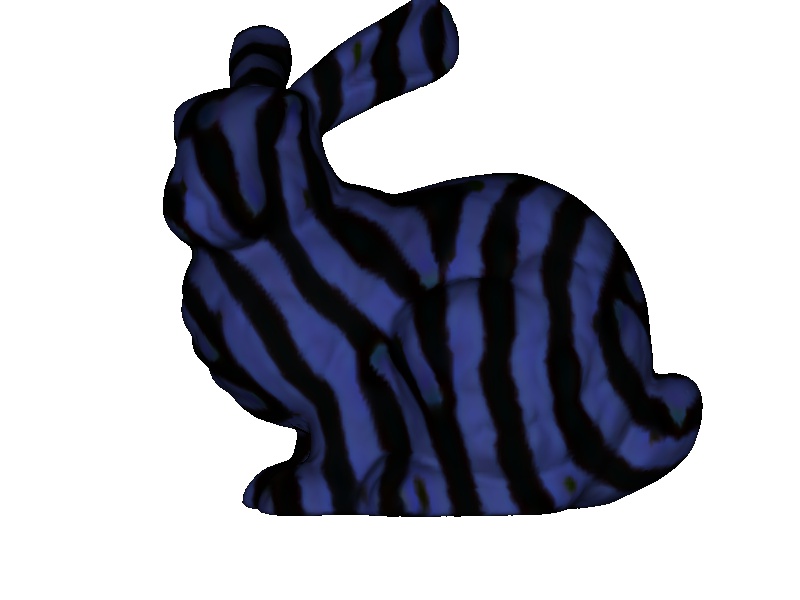}
\centering
\caption{RD texture models can easily be applied to mesh surfaces. We treat each vertex of the Stanford Bunny model as a cell, and allow the associated state vectors to diffuse along the mesh edges. This enables consistent texturing of dense meshes without constructing UV-maps or tangent coordinate frames. Please see the supplementary videos for the system dynamics and more views.}
\label{fig:surface}
\end{figure}

\begin{figure}[t!]
\includegraphics[width=4.0cm]{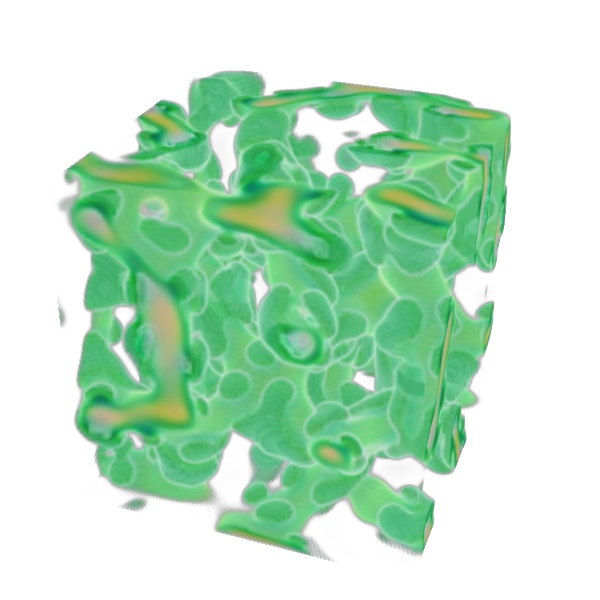}
\includegraphics[width=4.0cm]{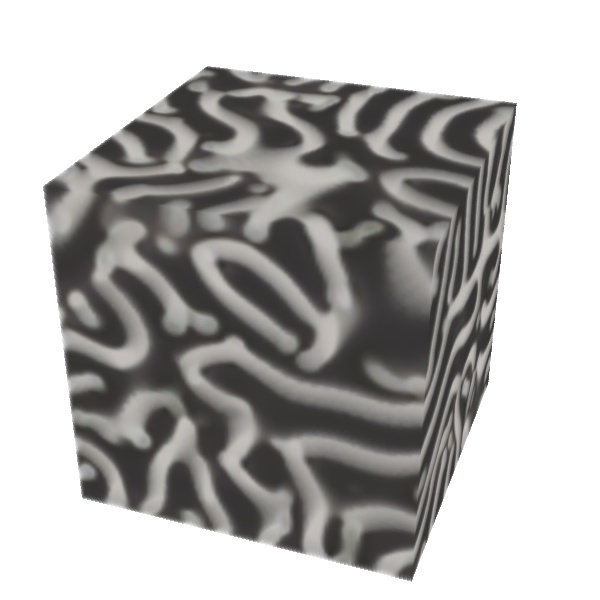}
\includegraphics[width=4.0cm]{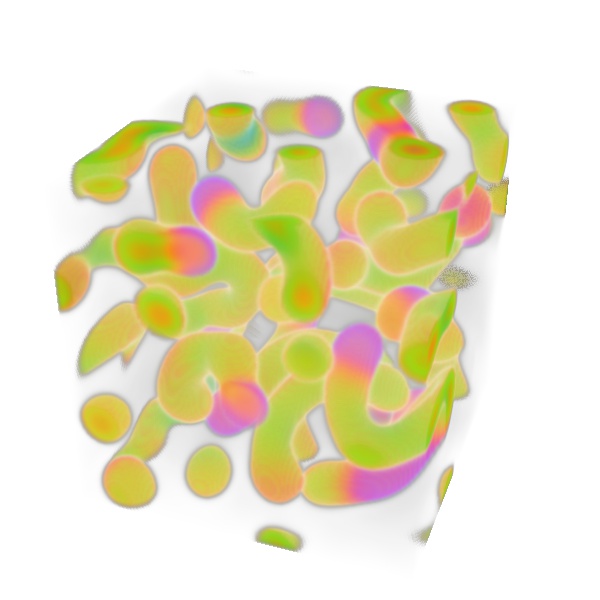}
\includegraphics[width=4.0cm]{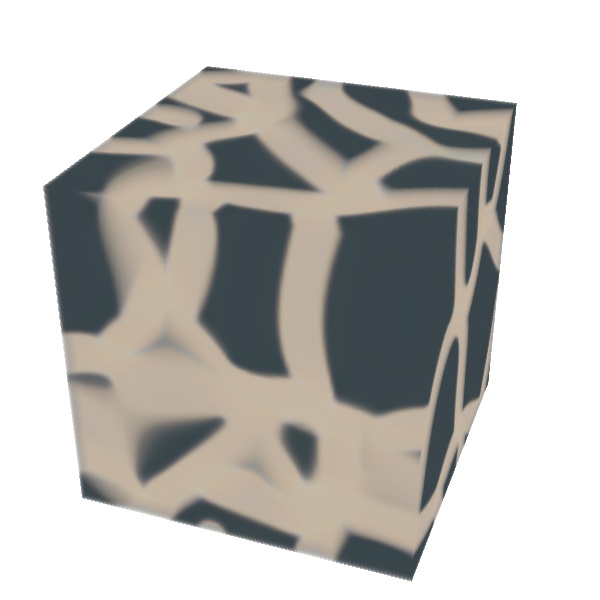}
\includegraphics[width=4.0cm]{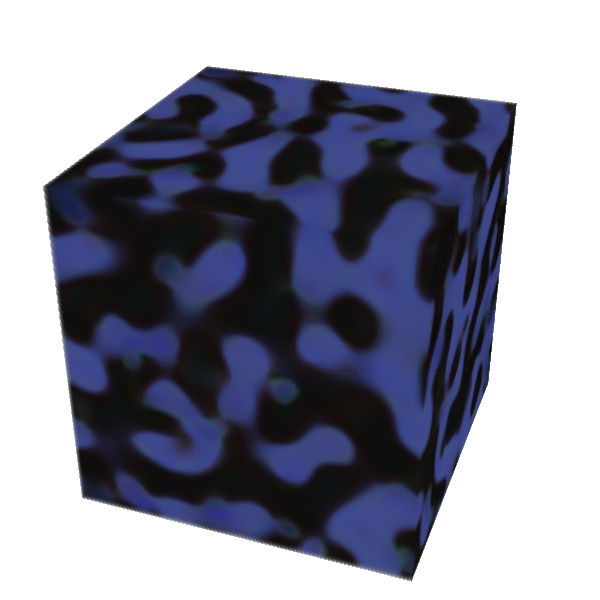}
\includegraphics[width=4.0cm]{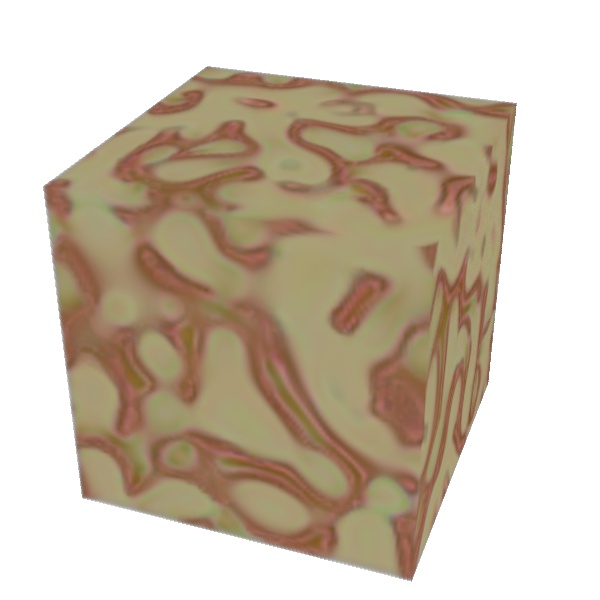}
\centering
\caption{2D to 3D generalization of Reaction-Diffusion models. The models that were trained on 2D plane with 2D image loss and executed on a 3D space. For some patterns, individual slices though the volume have textures similar to the target images, while for others (e.g. the last row) the similarity is less convincing. We treat white color as transparent to visualize the structure of {\tt lizards} and {\tt polka-dotted} patterns. Please see supplementary videos for model dynamics.}
\label{fig:volume}
\end{figure}

\paragraph{Generalization beyond 2D plane} RD-system can be applied to any manifold that has a diffusion operation defined on it. This enables much more extreme out-of-training generalization scenarios, than those possible for Neural CA. For example, applying texture synthesis method by \cite{niklasson2021self-organising} to a 3d mesh surface would require defining smooth local tangent coordinate systems for all surface cells (for example, located at mesh vertices). This is necessary to compute partial derivatives of cell states with some variant of generalized of Sobel filters. In contrast, Neural RD doesn't require tangents, and can be applied to an arbitrary mesh by state diffusion over the edges of the mesh graph (see fig. \ref{fig:surface}). Even more surprisingly, RD models that were trained to produce patterns on a 2d plane, can be applied to spaces of higher dimensionality by just replacing the diffusion operator with a 3d equivalent. Figure \ref{fig:volume} shows examples of volumetric texture synthesis by 2d models.

\section{Discussion}

Reaction-Diffusion is known to be an important mechanism controlling many developmental processes in nature \citep{Kondo2010ReactionDiffusionMA, Landge2019PatternFM}. We think that mastering RD-system engineering is an important prerequisite for making human technology more life-like: robust, flexible, and sustainable. This work demonstrates the applicability of Differentiable Programming to the design of RD-systems. We think this is an important stepping stone to transform RD into a practical engineering tool.

To achieve this, some limitations should be addressed in future work. First, it is crucial to find such optimization problem formulations that would produce physically plausible RD systems. Second, further research in the area of differentiable objective formulations is needed to make this approach applicable to a broader range of design problems for self-organizing systems.


\footnotesize
\bibliographystyle{apalike}
\bibliography{example} 

\begin{thebibliography}{}

\bibitem[Chan, 2020]{Chan2020LeniaAE}
Chan, B. (2020).
\newblock Lenia and expanded universe.
\newblock {\em ArXiv}, abs/2005.03742.

\bibitem[Cimpoi et~al., 2014]{Cimpoi2014DescribingTI}
Cimpoi, M., Maji, S., Kokkinos, I., Mohamed, S., and Vedaldi, A. (2014).
\newblock Describing textures in the wild.
\newblock {\em 2014 IEEE Conference on Computer Vision and Pattern
  Recognition}, pages 3606--3613.

\bibitem[Gatys et~al., 2015]{Gatys2015TextureSU}
Gatys, L.~A., Ecker, A.~S., and Bethge, M. (2015).
\newblock Texture synthesis using convolutional neural networks.
\newblock In {\em NIPS}.

\bibitem[Kingma and Ba, 2015]{Kingma2015AdamAM}
Kingma, D.~P. and Ba, J. (2015).
\newblock Adam: A method for stochastic optimization.
\newblock {\em CoRR}, abs/1412.6980.

\bibitem[Kondo and Miura, 2010]{Kondo2010ReactionDiffusionMA}
Kondo, S. and Miura, T. (2010).
\newblock Reaction-diffusion model as a framework for understanding biological
  pattern formation.
\newblock {\em Science}, 329:1616 -- 1620.

\bibitem[Landge et~al., 2019]{Landge2019PatternFM}
Landge, A.~N., Jordan, B.~M., Diego, X., and M{\"u}ller, P. (2019).
\newblock Pattern formation mechanisms of self-organizing reaction-diffusion
  systems.
\newblock {\em Developmental Biology}, 460:2 -- 11.

\bibitem[Mordvintsev et~al., 2020]{mordvintsev2020growing}
Mordvintsev, A., Randazzo, E., Niklasson, E., and Levin, M. (2020).
\newblock Growing neural cellular automata.
\newblock {\em Distill}.
\newblock https://distill.pub/2020/growing-ca.

\bibitem[Niklasson et~al., 2021]{niklasson2021self-organising}
Niklasson, E., Mordvintsev, A., Randazzo, E., and Levin, M. (2021).
\newblock Self-organising textures.
\newblock {\em Distill}.
\newblock https://distill.pub/selforg/2021/textures.

\bibitem[Pearson, 1993]{Pearson_1993}
Pearson, J.~E. (1993).
\newblock Complex patterns in a simple system.
\newblock {\em Science}, 261(5118):189–192.

\bibitem[Ramachandran et~al., 2017]{Ramachandran2017SwishAS}
Ramachandran, P., Zoph, B., and Le, Q.~V. (2017).
\newblock Swish: a self-gated activation function.
\newblock {\em arXiv: Neural and Evolutionary Computing}.

\bibitem[Randazzo et~al., 2020]{randazzo2020self-classifying}
Randazzo, E., Mordvintsev, A., Niklasson, E., Levin, M., and Greydanus, S.
  (2020).
\newblock Self-classifying mnist digits.
\newblock {\em Distill}.
\newblock https://distill.pub/2020/selforg/mnist.

\bibitem[Scalise and Schulman, 2014]{Scalise2014DesigningMR}
Scalise, D. and Schulman, R. (2014).
\newblock Designing modular reaction-diffusion programs for complex pattern
  formation.

\bibitem[Turing, 1952]{Turing_1952}
Turing, A.~M. (1952).
\newblock The chemical basis of morphogenesis.
\newblock {\em Philosophical transactions of the Royal Society of London.
  Series B, Biological sciences}, 237(641):37–72.

\bibitem[Turk, 1991]{Turk1991GeneratingTO}
Turk, G. (1991).
\newblock Generating textures on arbitrary surfaces using reaction-diffusion.
\newblock {\em Proceedings of the 18th annual conference on Computer graphics
  and interactive techniques}.

\bibitem[Witkin and Kass, 1991]{Witkin1991ReactiondiffusionT}
Witkin, A. and Kass, M. (1991).
\newblock Reaction-diffusion textures.
\newblock {\em Proceedings of the 18th annual conference on Computer graphics
  and interactive techniques}.

\end{thebibliography}

\end{document}